\documentclass[10pt,twocolumn,letterpaper]{article}

\usepackage[pagenumbers]{cvpr} 

\usepackage{graphicx}
\usepackage{amsmath}
\usepackage{amssymb}

\usepackage{xcolor}
\usepackage{enumitem}
\usepackage{xspace}
\usepackage{booktabs}
\usepackage{colortbl}
\usepackage{multirow}
\usepackage{makecell}
\usepackage{lipsum} 
\usepackage{gensymb} 
\usepackage{mathtools}
\usepackage[accsupp]{axessibility} 

\usepackage[labelsep=period]{caption}
\renewcommand{\paragraph}[1]{\vspace{0.2em}\noindent \textbf{#1 \hspace{0.2em}}}
\newcommand{\methodName}{UniPart\xspace}
\newcommand{\repName}{Geom-Seg~VecSet\xspace}

\usepackage[marginal]{footmisc}

\newcommand{\wlink}[1]{\textcolor{magenta}{{#1}}}

\definecolor{MyDarkRed}{rgb}{0.46, 0.16, 0.16}
\definecolor{MyDarkBlue}{rgb}{0.16, 0.16, 0.66}

\definecolor{cvprblue}{rgb}{0.21,0.49,0.74}
\usepackage[breaklinks,colorlinks,allcolors=cvprblue]{hyperref}

\def\paperID{5778} 
\def\confName{CVPR}
\def\confYear{2026}

\newcommand{\PaperTitleText}{UniPart: Part-Level 3D Generation with Unified 3D Geom-Seg Latents}
\title{\PaperTitleText}

\newcommand{\PaperAuthorBlock}{
Xufan He$^{1,2}\footnotemark[1]$ \quad Yushuang Wu$^{2}\footnotemark[1]$ \quad Xiaoyang Guo$^{2}$ \quad Chongjie Ye$^{4,5}$  \quad Jiaqing Zhou$^{2}$ \\ \quad Tianlei Hu$^{3}$  \quad Xiaoguang Han$^{5,4,6}$ \quad Dong Du$^{1}\footnotemark[2]$ \\
{\normalsize $^1$Nanjing University of Science and Technology}
{\normalsize $^2$ByteDance Games}
{\normalsize $^3$Zhejiang University}\\
{\normalsize $^4$FNii-Shenzhen}
{\normalsize $^5$SSE, CUHKSZ}
{\normalsize $^6$Guangdong Provincial Key Laboratory of Future Networks of Intelligence}
}
\author{\PaperAuthorBlock}

\makeatletter
\let\SavedCVPRMaketitle\@maketitle
\makeatother

\begin{document}

\twocolumn[{%
\renewcommand\twocolumn[1][]{#1}%
\maketitle
\begin{center}
    \centering
    \captionsetup{type=figure}
    \vspace{-6mm}
    \includegraphics[width=1.0\textwidth]{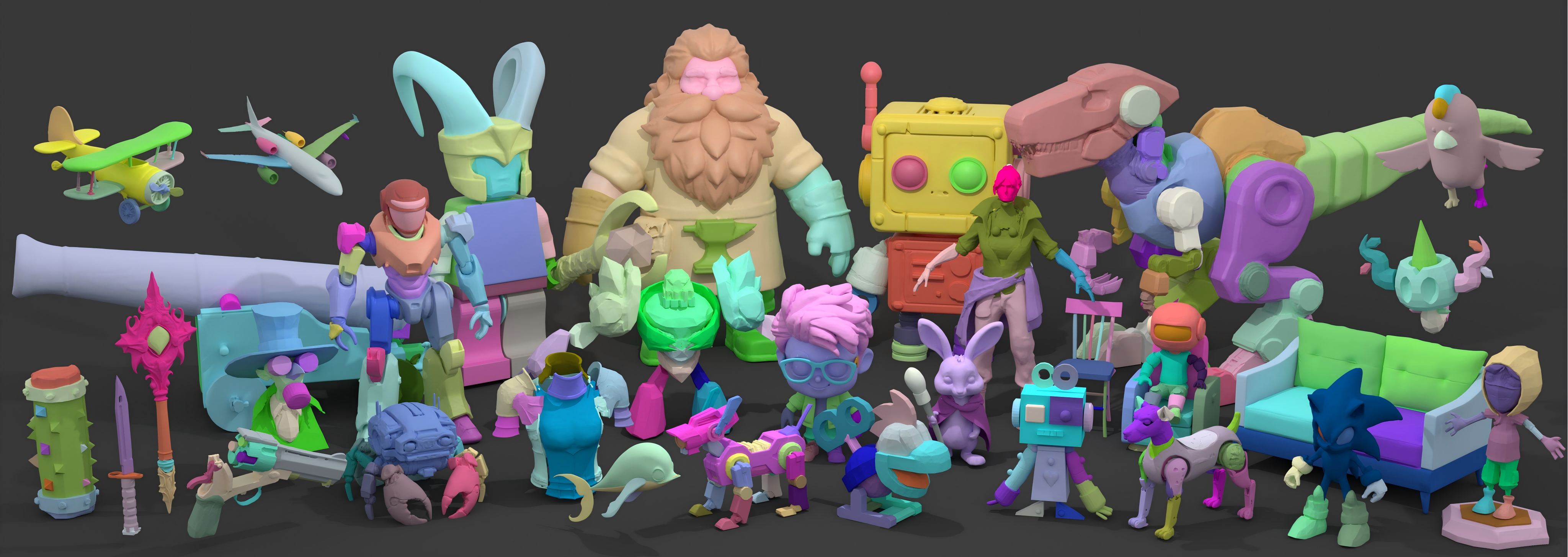}
    \vspace{-2mm}
    \label{fig:teaser}
\end{center}%
}]

\footnotetext[1]{Equal Contribution.}
\footnotetext[1]{This work was done by Xufan He as an intern at ByteDance, supervised by Yushuang Wu.}
\footnotetext[2]{Corresponding author: \wlink{dongdu@njust.edu.cn}.}

\begin{abstract}
Part-level 3D generation is essential for applications requiring decomposable and structured 3D synthesis. However, existing methods either rely on implicit part segmentation with limited granularity control or depend on strong external segmenters trained on large annotated datasets. In this work, we observe that part awareness emerges naturally during whole-object geometry learning and propose \repName, a unified geometry-segmentation latent representation that jointly encodes object geometry and part-level structure. Building on this representation, we introduce \methodName, a two-stage latent diffusion framework for image-guided part-level 3D generation. The first stage performs joint geometry generation and latent part segmentation, while the second stage conditions part-level diffusion on both whole-object and part-specific latents. A dual-space generation scheme further enhances geometric fidelity by predicting part latents in both global and canonical spaces. Extensive experiments demonstrate that \methodName achieves superior segmentation controllability and part-level geometric quality compared with existing approaches.
\end{abstract}
\vspace{-7mm}

\section{Introduction}
\label{sec:intro}
With the rapid development of 3D content creation in fields such as computer graphics, robotics, virtual reality, and digital manufacturing, the demand for controllable and semantically structured 3D shape generation has increased significantly. Traditional 3D generation methods~\cite{zhang2024clay,zhao2025hunyuan3d2,li2025triposg,xiang2025structured} primarily focus on producing holistic meshes that capture only the overall geometry of objects. However, real-world applications such as part-aware editing, physical simulation, robotic manipulation, and modular design—require not only accurate geometry but also a decomposable understanding of object parts and their relationships. This need has motivated increasing interest in part-level 3D shape generation, which aims to synthesize objects as sets of coherent and semantically meaningful components.
Compared to holistic 3D reconstruction, part-level 3D shape generation poses two principal challenges. First, achieving structurally and semantically coherent part segmentation is inherently challenging. The model must infer an appropriate segmentation granularity while preserving both global structural integrity and semantic consistency.  The wide variability in object geometries and part compositions across and even within categories further makes coherent and meaningful decomposition highly nontrivial. Second, part geometry degradation frequently occurs during generation, particularly for small or thin components. These fine structures tend to lose geometric fidelity due to limited generative resolution and the difficulty of jointly preserving global coherence and local geometric details.

\begin{figure}[t]  %
    \centering
    \captionsetup{type=figure}
    \includegraphics[width=\linewidth]{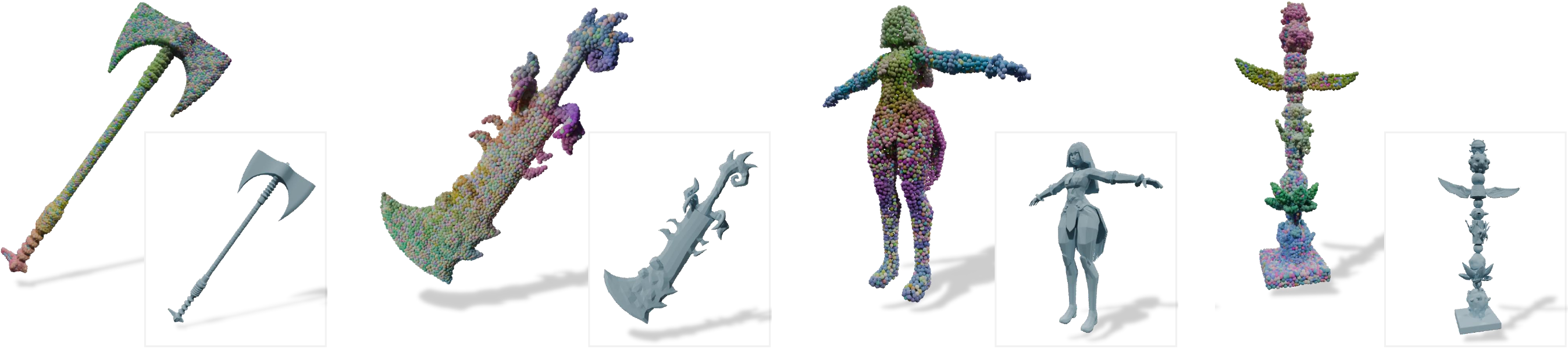}
    \caption{
        Latent correlation maps from Hunyuan3D-2.1 DiT attention at inference. Latents (points) correlate strongly within the same semantic part, suggesting implicit part awareness.
        }
    \label{fig:attn_map}
\end{figure}

Existing solutions either perform implicit segmentation by grouping latent vectors~\cite{lin2025partcrafter, tang2025partpacker, chen2025autopartgen} or follow a generate-then-segment paradigm with explicit representations such as multi-view images~\cite{chen2025partgen}, dense points~\cite{yan2025xpart}, low-resolution voxels~\cite{yang2025omnipart}, and mesh faces~\cite{yang2025holopart}. The former often provides limited control over segmentation granularity and part-level fidelity, while the latter relies on an auxiliary segmenter trained with substantial part annotations. In this work, we start from a different perspective: \textbf{part awareness can emerge implicitly during whole-geometry learning}.
To provide evidence for this perspective, we visualize latent correlation maps from self-attention in Hunyuan3D-2.1 DiT during inference. As shown in Fig.~\ref{fig:attn_map}, latents represented as points exhibit stronger correlations within the same object part, indicating that part-aware structure is already formed in pure whole-object geometry generation. 
Motivated by this observation, we construct a unified 3D geometry-segmentation latent space based on VecSet~\cite{zhang2024clay}. The proposed representation, \repName, jointly encodes geometry and part semantics, and decodes each latent vector into both its geometric contribution and part label. By coupling part understanding with geometry modeling in a single latent space, \repName preserves reconstruction quality comparable to the original VecSet while enabling explicit part reasoning.

For image-guided part-level 3D generation, we introduce a two-stage latent diffusion model, \methodName. We first perform whole-object diffusion to generate \repName, yielding both the global geometric latent and the corresponding part latent mask. Compared with methods that conduct part-level generation with implicit segmentation, our approach produces explicit part-latent masks, providing more flexible segmentation granularity and higher part-level controllability. Unlike previous generate-then-segment pipelines, our approach performs geometry generation and part segmentation jointly. This enables more efficient part learning by leveraging the rich part-aware priors implicitly captured in the geometry generator. Moreover, latent-space segmentation actually strikes a balance between granularity-akin to dense point-based approaches-and robustness-similar to low-resolution voxel methods. In the second stage, we directly use the segmented part latents, along with the global latent, as dual conditioning signals for part-level generation. Compared with prior approaches that require whole-object geometry generation and segmentation on explicit representations to condition the part-level diffusion~\cite{yan2025xpart,yang2025holopart,yang2025omnipart}, our latent-based conditioning offers three key advantages: (i) avoiding geometry degradation introduced by decode-encode cycles (\eg, decoding to SDF for lossy marching-cube mesh extraction and re-encoding sampled surface points~\cite{yan2025xpart, yang2025holopart}); (ii) providing domain-aligned conditioning signals that facilitate latent-space diffusion learning; (iii) injecting informative part-whole relational cues inherently preserved in the latent space. Furthermore, our part-level diffusion performs a novel dual-space latent generation procedure that jointly predicts part geometry latents in both the global coordinate space and a normalized canonical space. Decoding these two latents provides the part’s canonical geometry (from the normalized space latent) and its scale and placement within the object (from the global-space latent), which substantially mitigates part-geometry degradation and enables seamless and consistent part composition. Extensive experiments are conducted to validate the effectiveness and superiority of our \methodName on both part segmentation and geometry quality. 

\paragraph{Contributions} Our key contributions are as follows:  
\begin{itemize}[itemsep=0pt,parsep=0pt,topsep=2bp]
    \item We introduce \repName, a unified geometry-segmentation latent representation that enables efficient part-level understanding directly from whole-object generative learning. 
    \item We develop \methodName, a two-stage latent diffusion model that first performs geometry generation and part-latent segmentation, and then conducts dual-space part diffusion for seamless part composition and full-resolution part geometry, conditioned on both part and whole latents.
    \item We conduct extensive experiments demonstrating the superiority of our method for part-level 3D object generation.
\end{itemize}

\section{Related Work}
\label{sec:related_works}

\paragraph{Object-level 3D Generation} 
Recent object-level 3D generation methods fall into two main categories: 2D-lifting and native 3D approaches. DreamFusion~\cite{poole2022dreamfusion} pioneered the use of 2D diffusion priors via Score Distillation Sampling~\cite{alldieck2024sds,wang2023sjc} to optimize 3D representations, while follow-ups~\cite{long2023wonder3d,wang2024prolificdreamer,liusyncdreamer,shi2023mvdream} improve multi-view consistency by generating aligned image sets before reconstruction. In parallel, native 3D generative models are trained directly on large-scale datasets~\cite{objaverse,deitke2024objaversexl}. CLAY~\cite{zhang2024clay} and related works~\cite{zhao2025hunyuan3d2,li2025triposg,li2024craftsman} adopt a VecSet-based VAE-LDM framework~\cite{zhang20233dshape2vecset} to generate meshes and require high-quality watertight meshes for training. Trellis~\cite{xiang2025structured} introduces an explicit voxel-based latent space trained with rendering losses, offering stronger geometric inductive bias~\cite{li2025sparc3d, he2025sparseflex}. However, scaling such representations to high resolution remains a key challenge for fine-detail synthesis. Our work operates at the part level and builds upon the VecSet paradigm: we retain the VecSet-based latent architecture but extend the purely geometric latent space by incorporating geometry-aligned segmentation features, thereby constructing a novel unified geometry-segmentation latent space that enables structured part decomposition while preserving reconstruction fidelity.

\paragraph{3D Part Segmentation} 
Recent part segmentation methods leverage either 2D priors or native 3D supervision. Approaches such as SaMesh~\cite{tang2024samesh} lift 2D SAM~\cite{kirillov2023sam, ravi2024sam2} predictions to 3D via multi-view post-processing, while SAMPart3D~\cite{yang2024sampart3d} and PartField~\cite{liu2025partfield} distill SAM priors into a 3D semantic feature field for clustering-based decomposition. In contrast, 3D native methods like P3-SAM~\cite{ma2025p3sam} and PartSAM~\cite{zhu2025partsam} are trained directly on large-scale part-annotated meshes and produce point-wise semantic features for mask prompting. GeoSAM2~\cite{deng2025geosam2} adopts a hybrid strategy: it finetunes the pretrained SAM on 3D data to enable multi-view consistent segmentation, then back-projects the 2D segmentation onto the mesh surface to obtain a coherent 3D part decomposition.
Unlike these strategies that conduct segmentation on explicit 2D/3D representations, our method performs decomposition on the latent representation. By integrating high-dimensional, geometry-aligned semantic features into the latent representation, our approach inherits the robustness and boundary-awareness of sampling while enabling end-to-end part-level generation without relying on hard segmentation masks.
Meanwhile, solving the segmentation task as a diffusion-sampling task is much more reasonable than regression. Unlike regression, which forces a single deterministic output, treating part segmentation as a diffusion-based sampling task better captures the inherent ambiguity and multi-modality of valid part decompositions.

\paragraph{Part-level 3D Generation} 
Existing part-level 3D generation methods mainly fall into two groups based on the diffusion and autoregressive paradigms. Diffusion-based approaches can be further categorized by their generation strategy: one-shot-for-all methods such as PartCrafter~\cite{lin2025partcrafter}, PartPacker~\cite{tang2025partpacker}, OmniPart~\cite{yang2025omnipart}, BANG~\cite{zhang2025bang}, FullPart~\cite{ding2025fullpart} and X-Part~\cite{yan2025xpart}, which jointly generate all parts in a single forward pass; one-shot-per-part models such as HoloPart~\cite{yang2025holopart} generates each part independently but conditioned on global context, and AutoPartGen~\cite{chen2025autopartgen} adopts a sequential generation scheme. In contrast, AR-based methods operate directly on mesh sequences: MeshCoder~\cite{dai2025meshcoderllmpoweredstructuredmesh} leverages an LLM to produce Blender scripts that are executed to reconstruct parts, and Mesh Silksong~\cite{song2025meshsilksongautoregressivemesh} employs a manifold-aware mesh representation for autoregressive part synthesis. 
Except for adopting joint geometry-segmentation diffusion at the first stage, our part-level diffusion is also different in: (i) using latent-based conditions rather than explicit representations, which avoids geometry degradation, provides domain-aligned conditioning signals, and also injects informative part-whole relational cues; (ii) performing a dual-space latent generation procedure that achieves full-resolution part geometry of high quality and enables seamless and consistent part composition. 

\section{Method}
\label{sec:method}

\begin{figure*}[t]
\centering
\includegraphics[width=1.0\linewidth]{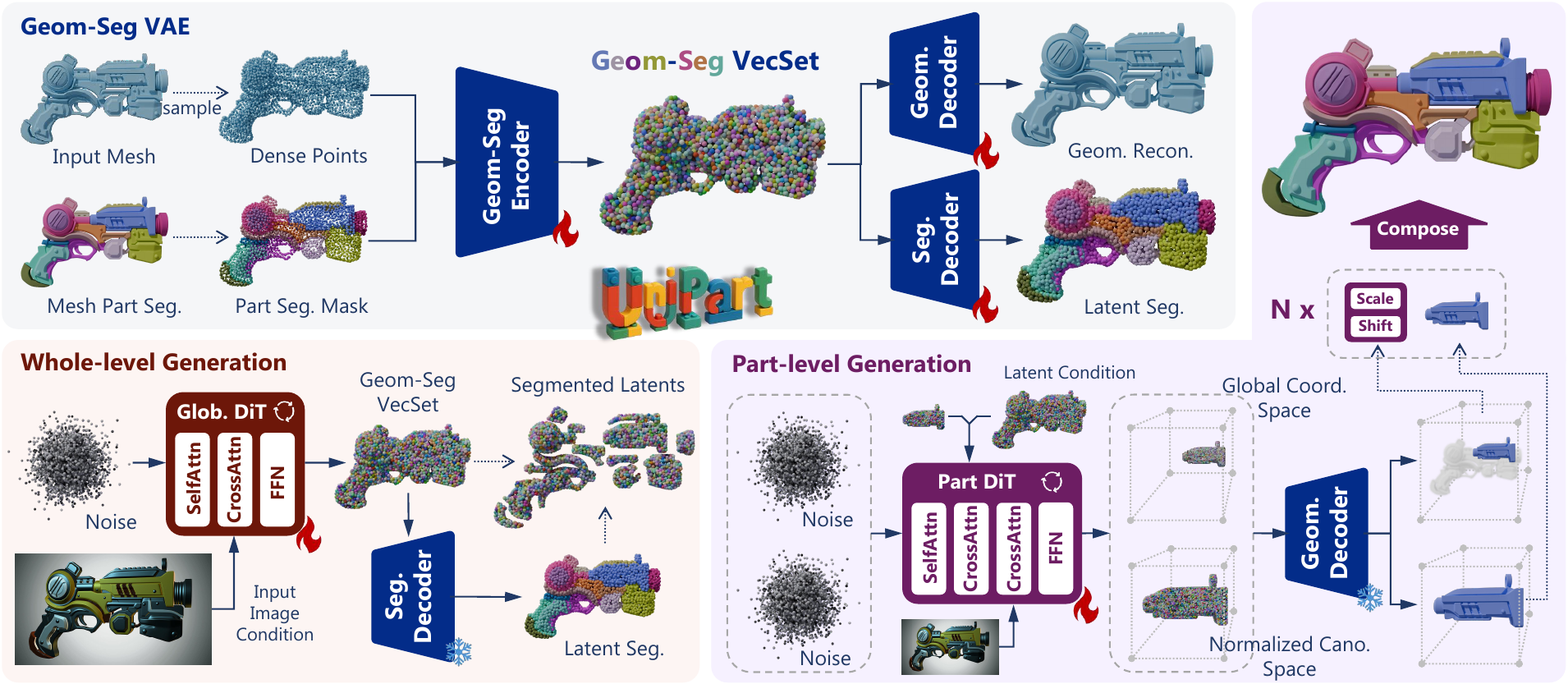}
\caption{The pipeline of \methodName. It includes a Geom-Seg VAE that encodes both whole geometry and part segmentation information into a unified representation, Geom-Seg VecSet. The image-guided part-level generation adopts a two-level pipeline, where a whole-level DiT first generates the whole geometry and segmented part latent, and a part-level DiT then accepts the input image and the whole-part latent as conditions for dual-space part latent generation. The final object mesh is composed of each full-resolution part mesh. }
\vspace{-3mm}
\label{fig:pipeline}
\end{figure*}

The task of part-level 3D shape generation aims to produce a decomposable 3D mesh $\mathcal{O}=\{\mathcal{M}_i\}_{i=1}^N$ from a prompt RGB image $I$, where each part of the object is represented by an individual mesh $\mathcal{M}$ and the number of parts $N$ is usually uncertain, depending on the complexity of the target object. The architecture of the proposed \methodName is illustrated in Fig.~\ref{fig:pipeline}. It consists of (i) a geometry-segmentation variational autoencoder that represents both the geometry and segmentation information of each object mesh in the latent space; (ii) a global-level diffusion model that generates the global geometry attached with part segmentation information in the latent space; and (iii) a part-level diffusion model that synthesizes the fine-grained part geometry by conditioning on each part latent segment and the global geometry latent. Finally, all individual parts compose the target object mesh.

\subsection{Geometry-Segmentation VAE}
\label{sec:vae}
\paragraph{Preliminary: VecSet}\cite{zhang2024clay} is a popular representation that encodes the geometry of a 3D mesh into the latent space. Specifically, dense points $P \in \mathbb{R}^{C\times 6}$ (each of $C$ points is represented by its position and normal vector on the mesh surface) are first sampled from the mesh $\mathcal{O}$. A fixed-length set of latent vectors $Z \in\mathbb{R}^{L \times d}$ is computed through a cross-attention layer between $P$ and its downsampled $L$ points (or a learnable query set) and several self-attention layers, encoded into $d$-dimensional latent vectors. This process is conducted by the encoder $\mathcal{E}$ of a transformer-based VAE. The decoder $\mathcal{D}$ then decodes $Z$ into an implicit field (\eg occupancy field or signed-distance field) for mesh reconstruction, as $\mathcal{D}(Z, q) \rightarrow [0,1]$, where $q$ is the set of query points in the target 3D space. The whole VAE is trained by minimizing the loss function formulated as 
\begin{equation}
\mathcal{L}_\text{vecset} = \mathcal{L}_\text{recon} + \lambda_\text{kl} \cdot \mathcal{L}_\text{kl},
\end{equation}
where $\mathcal{L}_\text{recon}$ is computed as $\| \mathcal{D}(Z, q) - f(q)\|_2^2$ with $f$ denoting the ground truth implicit field, $L_\text{kl}$ is the KL penalty for the divergence between $Z$ and Gaussian distribution, and $\lambda_\text{kl} > 0$ is a weighting factor. 

\paragraph{\repName}Classic VecSet representations encode rich geometric information into the latent space. To better support compositional generation, we further propose Geom-Seg VecSet that additionally encodes part segmentation information into the latent space and conducts latent segmenting in the decoding stage for part-wise representation. In doing so, we first segment object meshes into $N$ parts according to both geometric and semantic patterns. Then $C$ points with part segmentation label are sampled, denoted as $P \in \mathbb{R}^{C\times 7}$, as the input of encoding, deriving the encoded latent $Z = \mathcal{E}(P)\in\mathbb{R}^{L \times d}$. A decoder $\mathcal{D}_\text{geom}$ as in VecSet is adopted for geometry reconstruction from $Z$. We further develop a segmentation decoder $\mathcal{D}_\text{seg}$ and follow the promptable segmentation pipeline of SAM2~\cite{ravi2024sam2}, which accepts $Z$ and latent indices $q$ as input and generates segmentation masks corresponding to the prompt. 
The \repName VAE is trained by minimizing the loss function formulated as
\begin{equation}
\mathcal{L}_\text{vecset} = \mathcal{L}_\text{recon} + \mathcal{L}_\text{seg} + \lambda_\text{kl} \cdot \mathcal{L}_\text{kl},
\end{equation}
where $L_\text{seg}$ is the segmentation loss following SAM2~\cite{ravi2024sam2}. Considering geometric cues often delineate semantic boundaries, the geometric features and part segmentation patterns of objects are inherently interdependent. It critically facilitates coupled representation learning of geometry and segmentation information. In practice, we found that the \repName space can be smoothly learned by fine-tuning a pure geometric VecSet VAE, with the latent space and neural network kept in the original scale (equal in $L$ and $d$), which is sufficient to jointly represent the segmentation and geometry without loss of quality (see visualizations in Sec.~\ref{sec:ablation}). 
\subsection{Whole-level Latent Diffusion}
\label{sec:globgen}
With the learned geometry-segmentation latent space, we propose a two-stage pipeline for decomposable part-level generation, which consists of a whole-level diffusion transformer (DiT)~\cite{peebles2023scalable} that generates the global geometry-segmentation latent and a part-level DiT to refine each part latent for complete part geometry of quality. 

\paragraph{Geom-Seg Latent Diffusion}Our global-level DiT employs a rectified flow model~\cite{lipman2022flowmatching}, which uses a linear interpolation forward process $Z_t = (1-t)Z_0 + t \epsilon$ to construct the noisy \repName latent $Z_t$ from the data sample $Z_0$ and noise $\epsilon \in \mathbb{R}^{L \times d}$ at a timestep $t \in [0,1]$. In the backward process, our DiT parameterized by $\theta$ learns to model the timestep-dependent velocity field, $v(Z_t,t) = \delta_t Z_t$, representing the moving velocity from the noise to the data sample. $\theta$ is optimized on the conditional flow matching (CFM) objective~\cite{lipman2022flowmatching}:
\begin{equation}
\mathcal{L}_\text{cfm}(\theta) = \mathbb{E}_{t,Z_0,\epsilon}\|v_\theta(Z_t,t |I) - (\epsilon-Z_0) \|_2^2.
\end{equation}
At the inference stage, we start from a randomly initialized noise and iteratively predict the velocity as $t$ decreases from 1 to 0, which progressively denoises $Z_t$ and derives the final global latent prediction $\hat{Z_0}$. 

\paragraph{Latent Segmentation}With the generated global latent $\hat{Z_0}$, we adopt a frozen $\mathcal{D}_\text{seg}$ to decode the part mask for latent segmentation. It accepts $\hat{Z_0}$ and a dense prompt $r = [1,2,\cdots,L]$ as inputs to predict the part label $m_i, i\in\{j\}_{j=1}^L$ for each latent vector in $\hat{Z_0}$. Besides, we also train a small decoder to recover the anchor point position (the downsampled points at the encoding stage) for each \repName latent vector $p_i^\text{latent} = \mathcal{D}_\text{pos}(\hat{z_i})$. $\mathcal{D}_{pos}$ is trained on a well-learned \repName space with encoder frozen. Therefore, the training of $\mathcal{D}_\text{pos}$ only serves for position recovery without affecting the latent space construction. With the set of segmentation masks $\{m_i\}_{i=1}^L$ and latents' anchor positions $\{p^\text{latent}_i\}_{i=1}^L$, we first sample prompts via farthest point sampling (FPS) and then apply non-maximum suppression (NMS) to post-process the resulting masks, yielding a set of masks $\{m_i\}_{i=1}^N$ as the final latent segmentation result, which can be used to derive the segmented part latents $\{X_i\}_{i=1}^N$ where each $X_i$ is composed of uncertain number of $z_j$ from $\hat{Z_0}$. 

\subsection{Part-level Latent Diffusion}
\label{sec:partgen}
Given the generated whole-object latent $\hat{Z_0}$ and a set of segmented part latents $\{X_i\}_{i=1}^N$, we directly use them as condition signals to provide both global object geometry and local part information to synthesize high-quality part meshes.

\paragraph{Dual-Space Diffusion}Our part-level latent diffusion model adopts a dual-space generation architecture that jointly generates each part’s geometric latent from two spaces. For the $i$-th part, we denote its dual-space latent as $X_i^* \coloneqq (X_i^\text{gcs}, X_i^\text{ncs}) \in \mathbb{R}^{L \times 2d}$. Specifically, $X_i^\text{gcs}$ denotes the part latent in the \textit{global coordinate space}, encoding the part’s relative scale, position, and compositional relationship to the whole object. In contrast, $X_i^\text{ncs}$ denotes the part latent in a \textit{normalized canonical space}, capturing high-fidelity part geometry within a normalized $[0,1]^3$ coordinate frame and sharing the same resolution as the whole-object latent (see illustration in Fig.~\ref{fig:pipeline}). 
During dual-space diffusion training, a DiT parameterized by $\omega$ predicts $X_i^*$ for the $i$-th part, conditioned on the input image $I$, the part latent $X_i$ (padded to length $L$), and the whole-object latent $Z_0$ (replaced with $\hat{Z_0}$ at inference time). The prediction follows the same flow-matching formulation as in the whole-level latent diffusion, so we omit the detailed equations for brevity.
To enable the transformer to distinguish tokens from the two geometric contexts, we inject a learnable space embedding $e^s \in \mathbb{R}^{1 \times d}$, where $s \in \{\text{gcs}, \text{ncs}\}$ indexes the coordinate space. The corresponding space embedding is added to $X_i^\text{gcs}$ and $X_i^\text{ncs}$ with broadcasting before feeding them into the transformer. Furthermore, we also adopt a global-local attention mechanism inspired by previous work~\cite{chen2025ultra3d, yan2025xpart}: 
\begin{align}
    \text{Attn}_\text{local}=\text{Softmax}(\frac{\sigma_q(X_i^s)^\top\sigma_k({X_i^s})}{\sqrt{h}})\sigma_v(X_i^s),\\
    \text{Attn}_\text{global}=\text{Softmax}(\frac{\sigma_q(X_i^*)^\top\sigma_k(X_i^*)}{\sqrt{h}})\sigma_v(X_i^*),
\end{align}
where $s \in \{\text{gcs}, \text{ncs}\}$ indexes the coordinate space, $\sigma_{q/k/v}$ denotes the linear projections for query, key, and value, $h$ is the channel dimension of each attention head, and the diffusion timestep $t$ is omitted for simplicity. This global-local attention mechanism encourages joint attention to both whole-object and part-level geometry, leading to higher-quality part generation. In particular, $X_i^\text{gcs}$ shares the same coordinate system as the conditioning latents and can be viewed as a geometric completion of $X_i$, making it easier to predict. On the other hand, $X_i^\text{gcs}$ and $X_i^\text{ncs}$ are highly correlated, sharing rich part-level geometric information. These properties allow the coupled dual-space diffusion mechanism to learn high-quality geometry of each part in its own normalized canonical space more efficiently.

\paragraph{Part Decoding and Composition}
At the inference stage, we iteratively denoise two random noise conditioned on $\hat{Z_0}$ and $X_i$ to get the denoised part latents in two spaces $X_i^\text{gcs}$ and $X_i^\text{ncs}$ for the $i$-th part. Using a shared pre-trained geometry decoder, we decode the latents into $\mathcal{M}_i^\text{gcs}$ and $\mathcal{M}_i^\text{ncs}$, respectively. In the final step, the scale and position of the $i$-th part mesh can be computed from $\mathcal{M}_i^\text{gcs}$, which is utilized to transform $\mathcal{M}_i^\text{ncs}$ of high geometric quality into the global coordinate space for part mesh composition. 

\section{Experiments}
\label{sec:Experiments}
\subsection{Implementation Details}

\paragraph{Dataset Curation}
We curate a dataset of 300K objects with part-level segmentation by integrating multiple public sources~\cite{deitke2024objaversexl,objaverse,collins2022abodatasetbenchmarksrealworld}. We obtain part annotations through the following pipeline:
1) The part label is extracted through the connectivity of mesh, which is further filtered by reviewing explosion-view renderings, because connectivity-based segmentation is unreliable for scanned data; 2) We apply the winding number~\cite{Barill:FW:2018} for remeshing, effectively achieving hole filling and gap closure to ensure that the mesh is watertight; 3) The part labels of the remeshed faces are determined based on the nearest face labels from the original raw mesh. 

\paragraph{Experiment Setting}
Since there is no reliable correspondence between generated parts and ground-truth parts, we concatenate all parts into a single unified mesh for evaluation to enable a holistic assessment of geometric fidelity. 
We employ Chamfer Distance (CD), F-Score, and Intersection over Union (IoU) for quantitative evaluation. All meshes are normalized to $[-1, 1]$, and the best score across rotations of $0^\circ$, $90^\circ$, $180^\circ$, and $270^\circ$ is selected to ensure a pose-agnostic comparison. We implement five baseline methods for comparison: (i) PartPacker~\cite{tang2025partpacker}, (ii) PartCrafter~\cite{lin2025partcrafter}, (iii) OmniPart~\cite{yang2025omnipart}, (iv) HoloPart~\cite{yang2025holopart}, and (v) X-Part~\cite{yan2025xpart}. 
Among them, methods (i, ii, iii) accept single-image input for generation, the same as our approach. Methods (iv, v) require mesh input, so we generate the holistic object mesh using Hunyuan3D-2.1~\cite{hunyuan3d2025hunyuan3d21}, the same as ours, to enable fair comparison. Note that all baseline methods are implemented with the open-sourced version for evaluation.

\paragraph{Implementation Details}
For the Geom-Seg VAE and global-level shape generation, we build upon Hunyuan3D-2.1~\cite{hunyuan3d2025hunyuan3d21}, using classifier-free guidance (CFG) with a dropout rate of 0.1 and the AdamW optimizer with a weight decay of 0.01. For part-level generation, we initialize the weights from Hunyuan3D-2.1 and fine-tune with the same AdamW settings. All training is conducted on 8 NVIDIA A800 GPUs (80GB) with a batch size of 32. More implementation details are included in the supplementary materials.

\subsection{Experiment Results}
\begin{figure}[t]  %
    \centering
    \captionsetup{type=figure}
    \includegraphics[width=\linewidth]{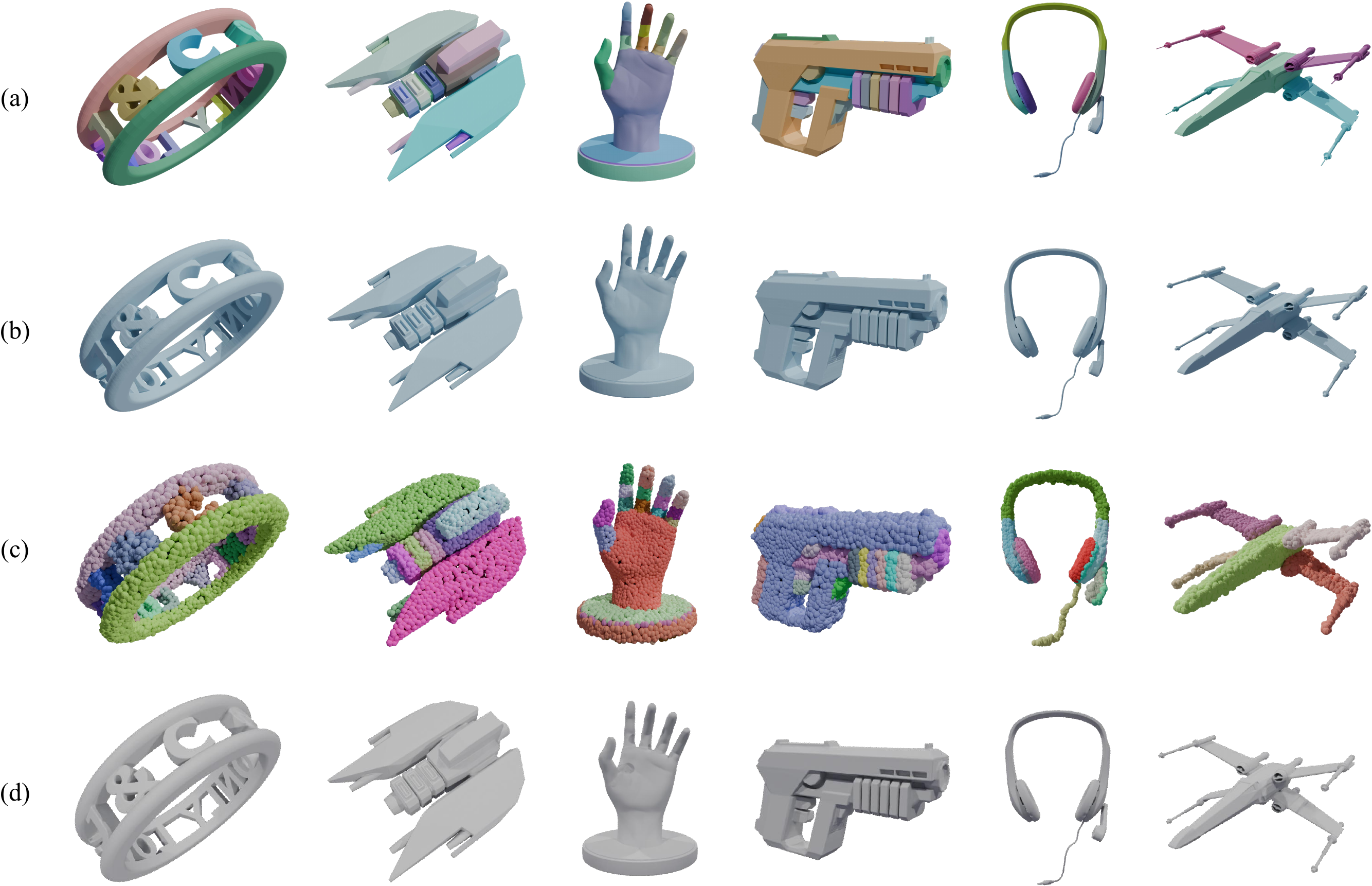}
    \caption{Reconstruction results of our Geom-Seg VAE. (a) Input mesh with part segmentation masks; (b) Reconstructed geometry; (c) Reconstructed part latent segmentation visualization. (d) Reconstructed geometry of Hunyuan3D-2.1~\cite{hunyuan3d2025hunyuan3d21} for reference. Please zoom in for details. }
    \label{fig:vae_recon}
    \vspace{-3mm}
\end{figure}
\paragraph{Geom-Seg VAE Reconstruction}
We first present the reconstruction performance of our proposed Geom-Seg VAE, as shown in Fig.~\ref{fig:vae_recon}. Both the reconstructed geometry and the inferred part-level latent segmentation demonstrate high fidelity to the input, preserving overall shape and part details while maintaining coherent part assignments across the object. We also provide results from Hunyuan3D-2.1~\cite{hunyuan3d2025hunyuan3d21} as a reference for the reconstructed geometry. These reference reconstructions offer a baseline for comparison, illustrating that our Geom-Seg VAE achieves accurate part-level segmentation while maintaining high-quality geometry. As an intermediate product, the whole-object geometry and part latent segmentation actually provide a solid support for the part-level diffusion at the second stage.

\begin{figure}[t]  %
    \centering
    \captionsetup{type=figure}
    \includegraphics[width=1.0\linewidth]{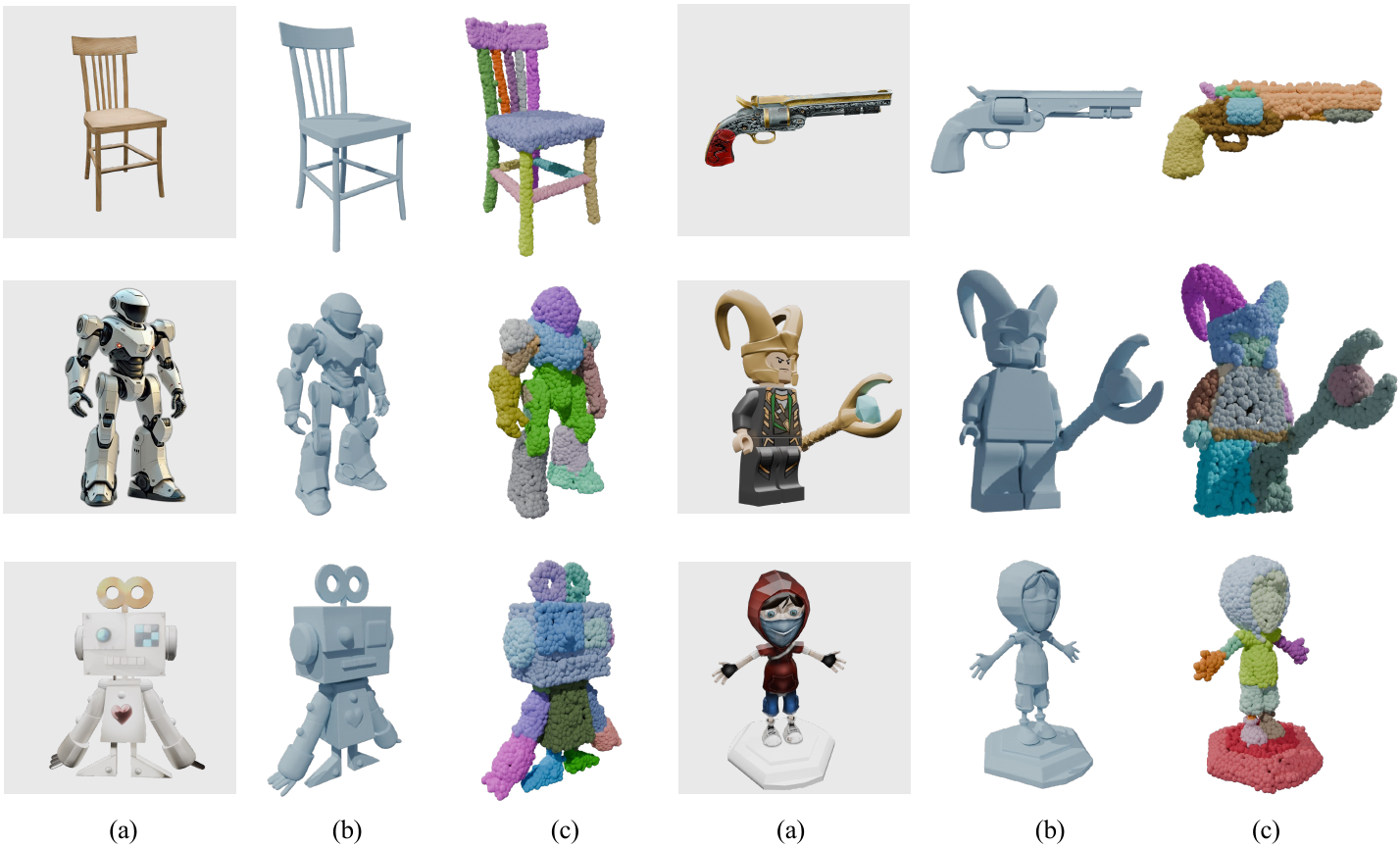}
    \caption{Generated results of our whole-level DiT. (a) Input image; (b) Generated whole-object geometry; (c) Generated part latent segmentation. Please zoom in for details. }
    \label{fig:seg_df}
\end{figure}

\paragraph{Whole-level Generation Results}
Fig.~\ref{fig:seg_df} visualizes some sampled outputs by the first-stage whole-level diffusion model to demonstrate its effectiveness. Accepting various types of object images as input, our whole-level DiT generates high-quality global geometry together with consistent part-level latent segmentations, indicating reliable synthesis of both object shape and part segmentation.

\begin{figure}
\centering
\includegraphics[width=1.\linewidth]{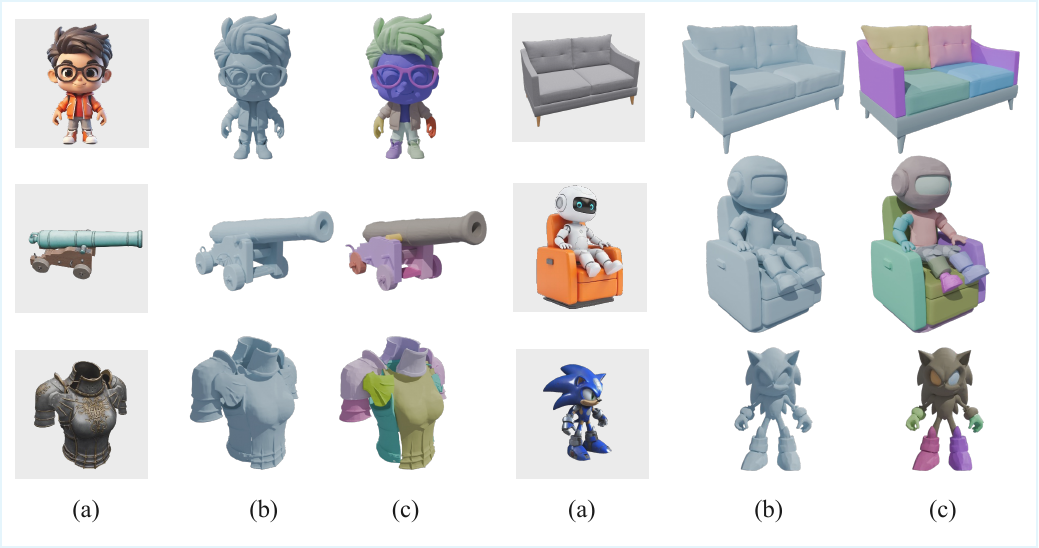}
\caption{More part-level generation results by our \methodName. (a) Input images; (b) The whole object geometry output by our whole-level DiT; (c) The composed part meshes output by our part-level DiT.}
\vspace{-3mm}
\label{fig:more_res}
\end{figure}

\begin{figure*}
\centering
\includegraphics[width=1.0\linewidth]{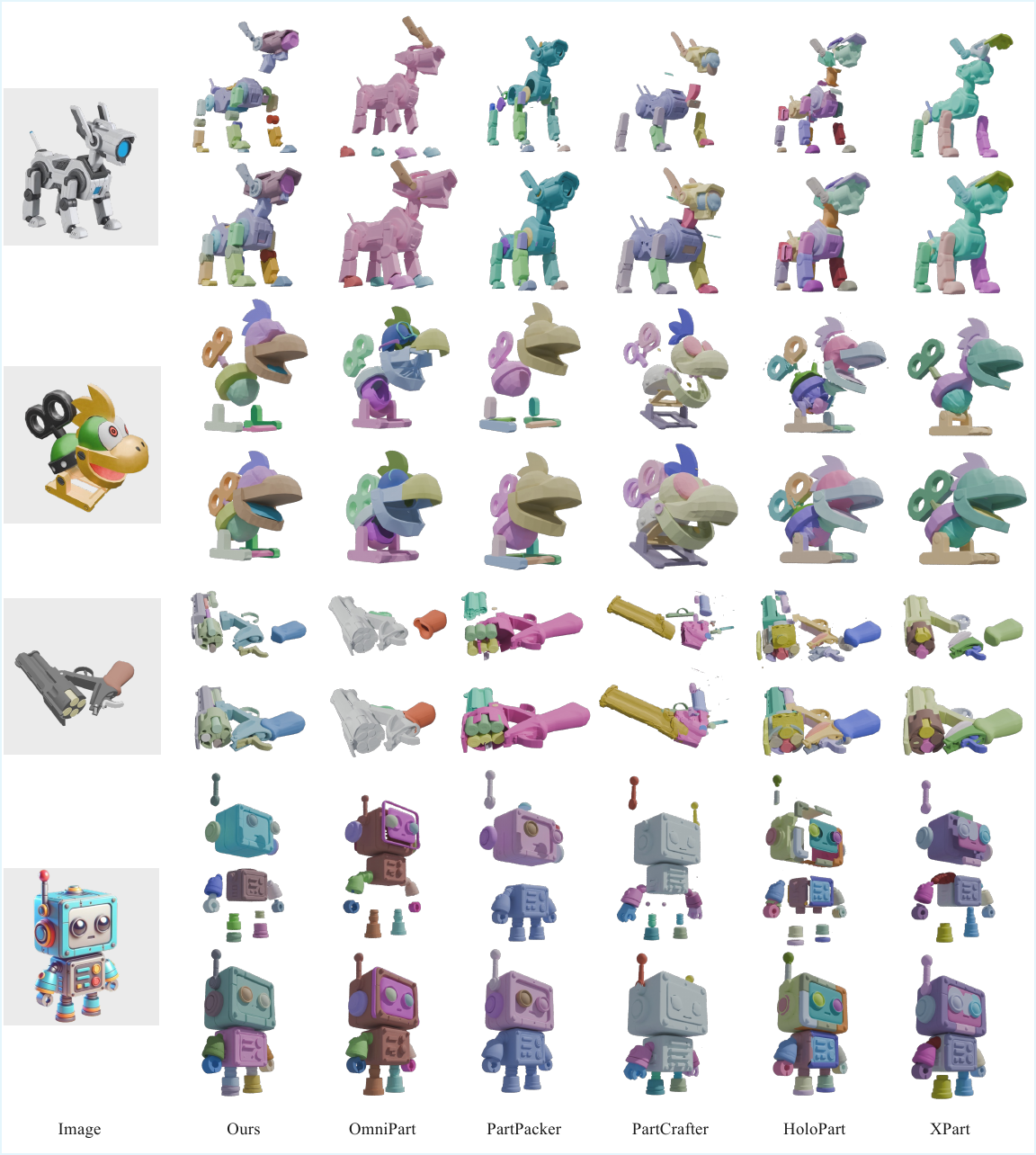}
\caption{Qualitative results of part-level 3D generation for a given image. We visualize the ``exploded'' parts in the first line of each mesh pair for better visualization of part generation. Our \methodName can produce more reasonable part segmentations and higher-quality part geometries.}
\vspace{-3mm}
\label{fig:part_results}
\end{figure*}

\paragraph{Part-level Generation Results}
We present both quantitative and qualitative evaluations of the final part-level generation produced by \methodName, and compare against baseline methods, including HoloPart~\cite{yang2025holopart}, OmniPart~\cite{yang2025omnipart}, PartCrafter~\cite{lin2025partcrafter}, PartPacker~\cite{tang2025partpacker}, and XPart~\cite{yan2025xpart}. Quantitative results are provided in Tab.~\ref{tab:abo_eval}. Our \methodName outperforms all other competitors consistently on all metrics (i.e., CD and F-Score with two distinct thresholds), which shows our method produces final objects with higher geometric quality and better alignment to the ground truth. Due to the space limitation, we include the quantitative segmentation comparison in the appendix. We further provide qualitative results in Fig.~\ref{fig:part_results}, which illustrates that our \methodName can generate more reasonable part segmentations and higher-quality part geometries.  
Some other part-level generations are visualized in Fig.~\ref{fig:more_res}. More visual results are shown in the teaser figure and supplementary materials. 





\begin{table}[htb]
\centering
\caption{Quantitative comparison on the geometry quality of part-level generation results. Best results are marked in bold font. Our \methodName significantly outperforms other methods on both CD and F-score metrics.}
\label{tab:abo_eval}
\resizebox{.45\textwidth}{!}{
\begin{tabular}{l|ccc}
\toprule
Method & CD$\downarrow$($\times10^2$) & F1@0.1$\uparrow$($\times10^2$) & IoU$\uparrow$($\times10^2$) \\
\midrule
HoloPart~\cite{yang2025holopart} & 2.86 & 82.33 & 23.40 \\
PartPacker~\cite{tang2025partpacker} & 2.18 & 74.35 & 13.97 \\
PartCrafter~\cite{lin2025partcrafter} & 1.03 & 46.08 & 5.63 \\
OmniPart~\cite{yang2025omnipart} & 1.99 & 85.04 & 27.94 \\
XPart~\cite{yan2025xpart} & 0.82 & 88.90 & \textbf{31.95} \\
UniPart (Ours) & \textbf{0.72} & \textbf{92.21} & 21.99 \\
\bottomrule
\end{tabular}
}
\vspace{-4mm}
\end{table}

\subsection{Ablation Studies}
\label{sec:ablation}
\paragraph{Module Design}
\begin{figure}
\centering
\includegraphics[width=0.94\linewidth]{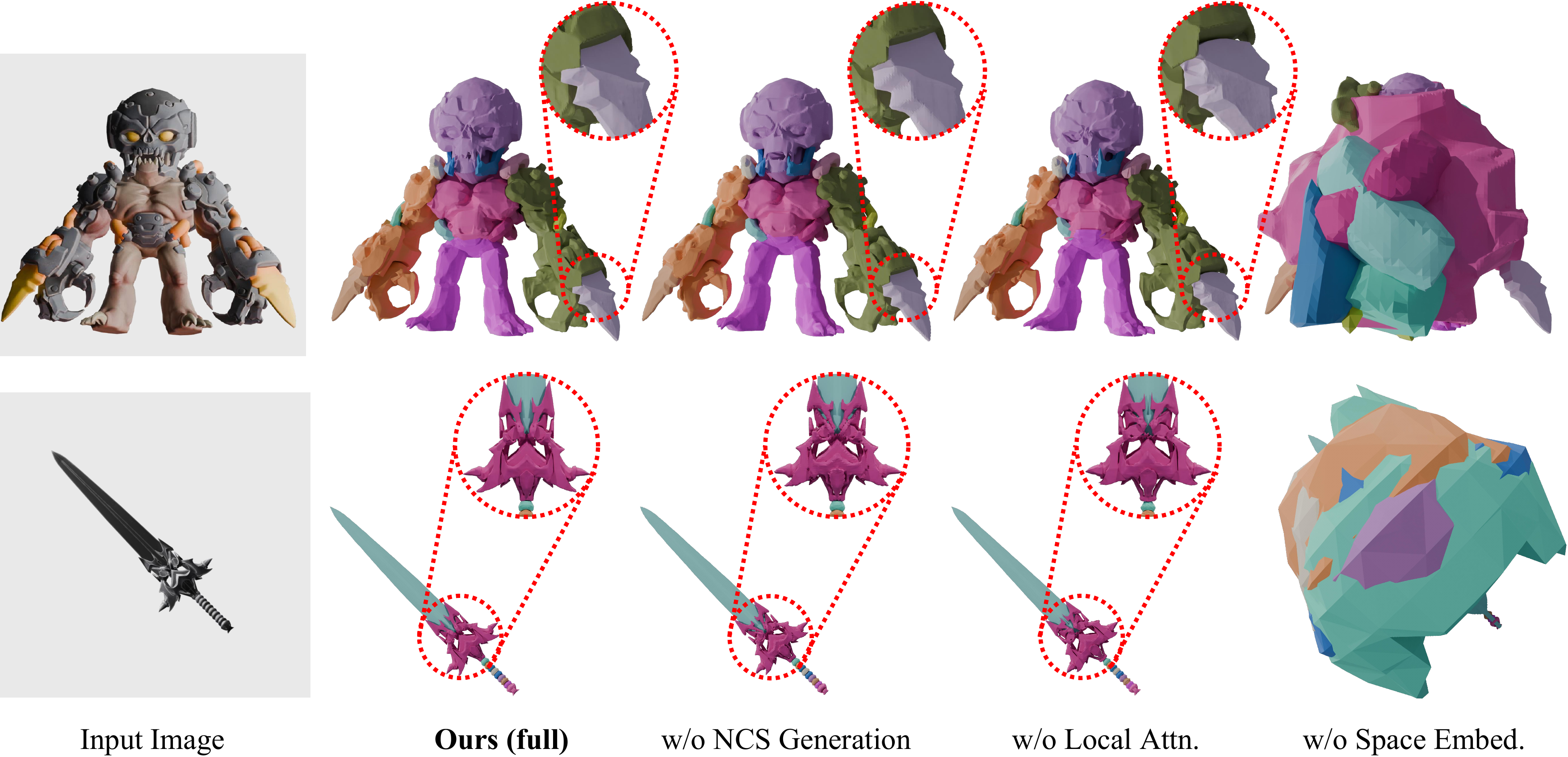}
\caption{Visual results of ablation studies for the design of \methodName, i.e., without using normalized canonical space (NCS) generation, local attention, and space embedding injection.}
\vspace{-4mm}
\label{fig:ablation}
\end{figure}
Comparison between the geometric reconstruction of \repName VAE and original VecSet VAE (Hunyuan3D-2.1 VAE) has been shown in Fig.~\ref{fig:vae_recon}, which proves that adding segmentation information into the latent space does not result in a drop in reconstruction quality. The learning of the whole-level DiT follows a typical latent diffusion model as in previous methods. Thus, we focus on the ablation study on the design of part-level DiT. There are three key designs: (i) dual-space generation for part meshes in the normalized canonical space (NCS), rather than directly using the global coordinate generation; (ii) local-global attention mechanism in the dual-space diffusion; (iii) space embedding injection for distinguishing tokens from different spaces within the attention mechanism to provide clearer guidance. We visualize the impact of each component using two examples shown in Fig.~\ref{fig:ablation}. 
The results validate that generation in the normalized canonical space effectively enhances part geometric fidelity, especially for small parts with fine details.
Local attention promotes part cohesion, yielding more harmonious spatial arrangements and improved global geometric consistency.
Without space embedding injection, the model frequently confuses the target space for each part, resulting in catastrophic failures during assembly.




\paragraph{Geom-Seg VecSet VAE}
The inclusion of additional segmentation information into VAE's latent space has almost no influence on the reconstruction quality, compared with the original Hunyuan3D-2.1 VAE, as illustrated in Tab.~\ref{tab:vae_comp}.

\begin{table}
\centering
\caption{Quantitative comparison on the geometry quality of part-level generation results. Best results are marked in bold font.}
\label{tab:vae_comp}
\resizebox{.38\textwidth}{!}{
\begin{tabular}{l|cc}
\toprule
Method & CD$\downarrow$($\times10^4$) & F1@0.01$\uparrow$($\times10^2$) \\
\midrule
TRELLIS\cite{xiang2025structured} & 1.32 & 80.59 \\
Dora\cite{Chen_2025_Dora} & 1.45 & 78.54 \\
Craftsman\cite{li2024craftsman} & 1.51 & 77.47 \\
XCubes\cite{ren2024xcube} & 1.42 & 77.57 \\
Hunyuan3D-2.1~\cite{hunyuan3d2025hunyuan3d21} & \textbf{1.29} & 80.85 \\
UniPart (Ours) & 1.30 & \textbf{80.89} \\
\bottomrule
\end{tabular}
}
\end{table}

\begin{figure}
\centering
\includegraphics[width=0.9\linewidth]{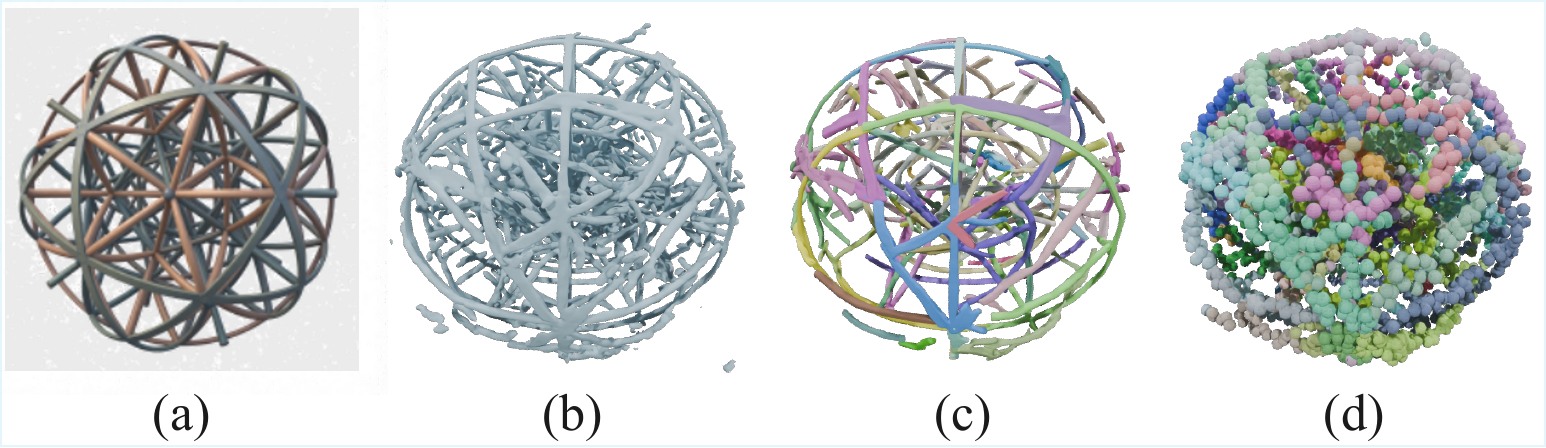}
\caption{Failure cases of our proposed \methodName. (a) Input images; (b) The whole object geometry output by our whole-level DiT; (c) The composed part meshes output by our part-level DiT; (d) The part latent segmentation visualization produced by our whole-level DiT.}
\vspace{-3mm}
\label{fig:failure_case}
\end{figure}

\paragraph{Failure Cases}
We also observe some failure cases, as shown in Fig.~\ref{fig:failure_case}. 
For inputs with high structural complexity, our joint generation of geometry and segmentation may fail, producing geometrically invalid or semantically inconsistent outputs.

\section{Conclusion}
\label{sec:Conclusion}
We propose a unified framework for part-level 3D generation, which uses a geometry-segmentation latent space and dual-stage latent diffusion. We build \repName, a joint latent representation encoding whole-object geometry and part structure, enabling part-aware understanding from whole-level generative learning without external segmenters. Our \methodName jointly generates geometry and performs part-level latent segmentation, then uses dual-space diffusion to predict part latents in global and canonical coordinates for high-fidelity, consistent part generation. Experiments show our method outperforms existing ones in controllability, fidelity, and generation quality, advancing structured 3D generation, editable 3D content creation, and part-aware generative modeling.


\section*{Acknowledgements}
This work was supported in part by the National Natural Science Foundation of China (No. 62502209), the Fundamental Research Funds for the Central Universities (No. 30925010538), the Basic Research Project No. HZQB-KCZYZ-2021067 of Hetao Shenzhen-HK S\&T Cooperation Zone, by Guangdong Provincial Outstanding Youth Fund with No. 2023B1515020055, the Shenzhen Outstanding Talents Training Fund 202002, the NSFC with Grant No. 62293482, the Guangdong Research Projects No. 2017ZT07X152 and No. 2019CX01X104, the Guangdong Provincial Key Laboratory of Future Networks of Intelligence (Grant No. 2022B1212010001), and the Shenzhen Key Laboratory of Big Data and Artificial Intelligence (Grant No. SYSPG20241211173853027), the Guangdong Province Radio Science Data Center.

\clearpage
\setcounter{section}{0}
\setcounter{subsection}{0}
\setcounter{subsubsection}{0}
\setcounter{table}{0}
\setcounter{figure}{0}
\renewcommand{\thesection}{S\arabic{section}}
\renewcommand{\theHsection}{supp.\arabic{section}}
\renewcommand{\thetable}{S\arabic{table}}
\renewcommand{\theHtable}{supp.\arabic{table}}
\renewcommand{\thefigure}{S\arabic{figure}}
\renewcommand{\theHfigure}{supp.\arabic{figure}}

\makeatletter
\gdef\@title{\PaperTitleText\\Supplementary Material}
\gdef\@author{\PaperAuthorBlock}
\gdef\@thanks{}
\makeatother

\twocolumn[{%
\renewcommand\twocolumn[1][]{#1}%
\makeatletter
\SavedCVPRMaketitle
\makeatother
}]
\footnotetext[1]{Equal Contribution.}
\footnotetext[1]{This work was done by Xufan He as an intern at ByteDance, supervised by Yushuang Wu.}
\footnotetext[2]{Corresponding author: \wlink{dongdu@njust.edu.cn}.}
\section{Implementation Details}
\paragraph{Network Architecture}
Our model is built upon the SOTA VectSet-based Hunyuan3D-2.1~\cite{hunyuan3d2025hunyuan3d21}. For the global-level DiT, we adopt a Transformer architecture aligned with Hunyuan-DiT~\cite{li2024hunyuandit}, enhanced with Mixture-of-Experts (MoE) layers to improve capacity and efficiency. Image conditions are encoded via DINOv2~\cite{oquab2023dinov2} and injected through cross-attention.
For the part-level DiT, we use the same architecture but condition on both the image and latent representations. The latent condition consists of segmented global latents, encoding global geometry and the target part mask. These latent vectors are processed by a lightweight 8-layer attention block and fused via cross-attention.
All conditions are randomly dropped with a 10\% probability to enable classifier-free guidance.

\paragraph{Segmentation Encoder}
We adapt the structure of VectSet~\cite{zhang20233dshape2vecset} encoder, append a part id embedding for each point after the position and normal. The part id embedding is drawn from a set of learnable embeddings, where the embedding corresponding to each part id is randomly selected, with the only constraint that different part ids correspond to distinct embedding vectors. Random selection avoids introducing spurious biases from predefined ID orderings, making the learned part representations more flexible and generalizable.

\paragraph{Segmentation Decoder}
We utilize a prompt-based segmentation mask decoder for generating an undetermined number of masks. Given a fixed number of latent tokens (e.g., 4096) and randomly sampled prompt tokens, a shallow MLP is adopted to predict whether each latent token belongs to the same part as the corresponding prompt token, thereby producing a segmentation mask.
To achieve automatic segmentation, we perform random dense sampling on the latent tokens to generate sufficient prompt candidates, ensuring full coverage of all parts. After obtaining a large number of initial masks, we apply non-maximum suppression (NMS) as post-processing to remove duplicate and redundant masks, retaining only the most reliable predictions. In this way, our decoder can flexibly and automatically produce high-quality segmentation results without predefining the number of output masks.

\paragraph{Position Decoder}
For dense mask post-processing and visualization, we train a position decoder to regress the precise 3D coordinate corresponding to each latent token. The feasibility of this design stems from the inherent spatial locality of the VectSet latent tokens. Although VectSet does not impose explicit spatial locality constraints on the latent representations, we observe that each latent token naturally possesses localized characteristics owing to its generation mechanism. Specifically, it first samples dense point clouds, followed by furthest point sampling (FPS) to extract a fixed number of representative points (e.g., 4096). These FPS-sampled points act as the queries in cross-attention layers, while the original dense point clouds serve as keys and values, yielding fixed-length latent tokens. The tokens are then mapped to the latent space via several stacked self-attention layers. As each latent token is derived from the cross-attention query initialized by an individual FPS point, it inherently encodes localized geometric information. Based on this property, the position decoder can reliably reconstruct the corresponding spatial position for every latent token. This locality is also the prerequisite for us to perform segmentation on the latent tokens.

\paragraph{Dataset Curation}
We construct a curated dataset by integrating multiple public sources~\cite{objaverseXL,objaverse,collins2022abodatasetbenchmarksrealworld}, yielding 300K objects with part-level segmentations.
Duplicate vertices are merged using a tolerance of $10^{-6}$, and meshes are split along connected components to isolate individual parts.
Parts with fewer than 6 faces or total face area below $10^{-3}$ are removed to eliminate small, noisy floating artifacts, and only meshes with part counts between 2 and 32 are retained to ensure meaningful structural diversity.
We filter out poorly segmented data, primarily from 3D scans, by manually reviewing explosion-view renderings, because connectivity-based segmentation is unreliable for scanned data.
We further apply the winding number~\cite{Barill:FW:2018} for remeshing, effectively achieving hole filling and gap closure to ensure that the mesh is watertight, and the part labels of the remeshed faces are determined based on the nearest face labels from the original raw mesh.

\section{Comparison}
\begin{figure*}
\centering
\includegraphics[width=0.8\linewidth]{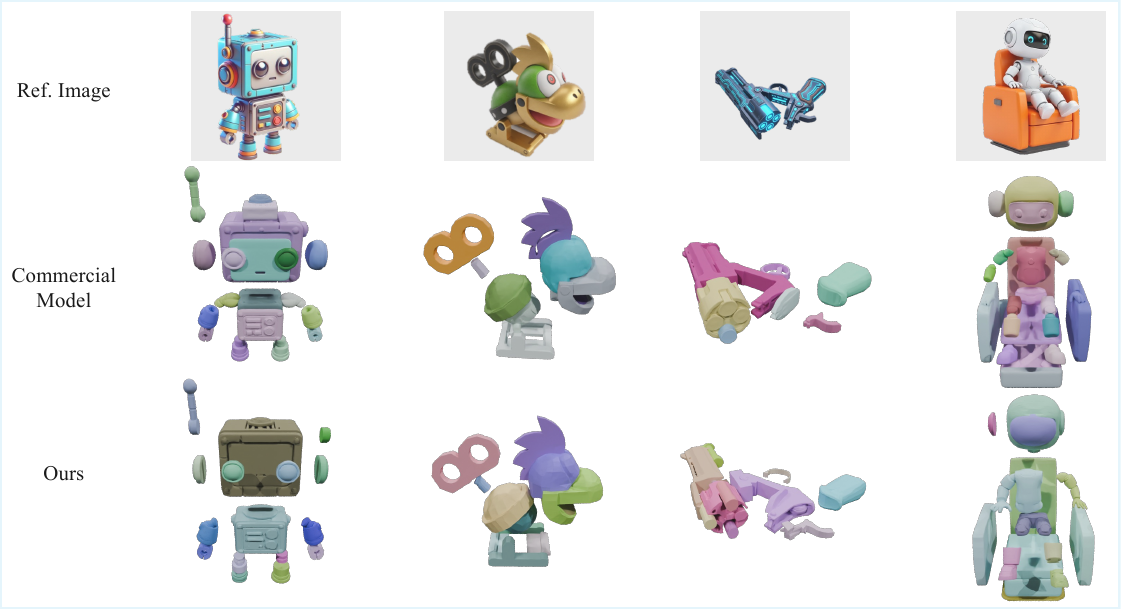}
\caption{Qualitative comparison with closed-source commercial models. We visualize the exploded views for better illustration of part-level generation. Our \methodName yields competitive results despite using a smaller-scale base model and less training data.}
\label{fig:comp_com}
\end{figure*}

\paragraph{Comparison with Closed-Source Commercial Models}
We additionally provide comparisons with closed-source commercial models. Some results are visualized in Fig.~\ref{fig:comp_com}. As shown, although with a smaller-scale base model for geometry generation and much less training data for part segmentation, our UniPart can still achieve comparable part-level generation results.

\paragraph{Comparison on 3D Segmentation}
Our UniPart actually conducts part segmentation during generation, so the segmentation performance can not be directly evaluated on existing 3D part segmentation benchmarks. Therefore, we specifically construct an evaluation set with 50 object meshes generated by our whole-level DiT (decode the global-geometry latent with Geom. decoder). A group of ten human annotators is invited to manually annotate the part segmentation on these meshes, also given the input image as a reference. The collected annotations are further double-checked and finely corrected by annotators in the second round to guarantee high quality. As a result, we obtain 536 high-quality part annotations from the 50 object meshes. Based on them, we evaluate the part segmentation results of UniPart with existing SOTA baseline methods, including SAMesh~\cite{tang2024samesh}, PartField~\cite{liu2025partfield}, and P3-SAM~\cite{ma2025p3sam}. For a direct comparison with ours, which only contains 4,096 latent vectors as part segmentation, we uniformly sample 4,096 points from each segmented point cloud or mesh surface of other methods to compute the numerical results. For quantitative results, we adopt mean IoU (mIoU) as the evaluation metric, with results shown in Tab.~\ref{tab:comp_numerical}.
\begin{table}[htb]
\centering
\caption{Quantitative comparison on the geometry quality of part-level generation results. Best results are marked in bold font.}
\label{tab:comp_numerical}
\resizebox{.48\textwidth}{!}{
\begin{tabular}{l|cccc}
\toprule
Method & SAMesh~\cite{tang2024samesh} & PartField~\cite{liu2025partfield} & P3-SAM~\cite{ma2025p3sam} & \textbf{UniPart (Ours) }  \\
\midrule
mIoU~$\uparrow$ & 0.3608 & 0.4167 & 0.7046 & \textbf{0.7222} \\
\bottomrule
\end{tabular}
}
\end{table}
\vspace{-3mm}

\section{Additional Results}
\begin{figure*}
\centering
\includegraphics[width=0.95\linewidth]{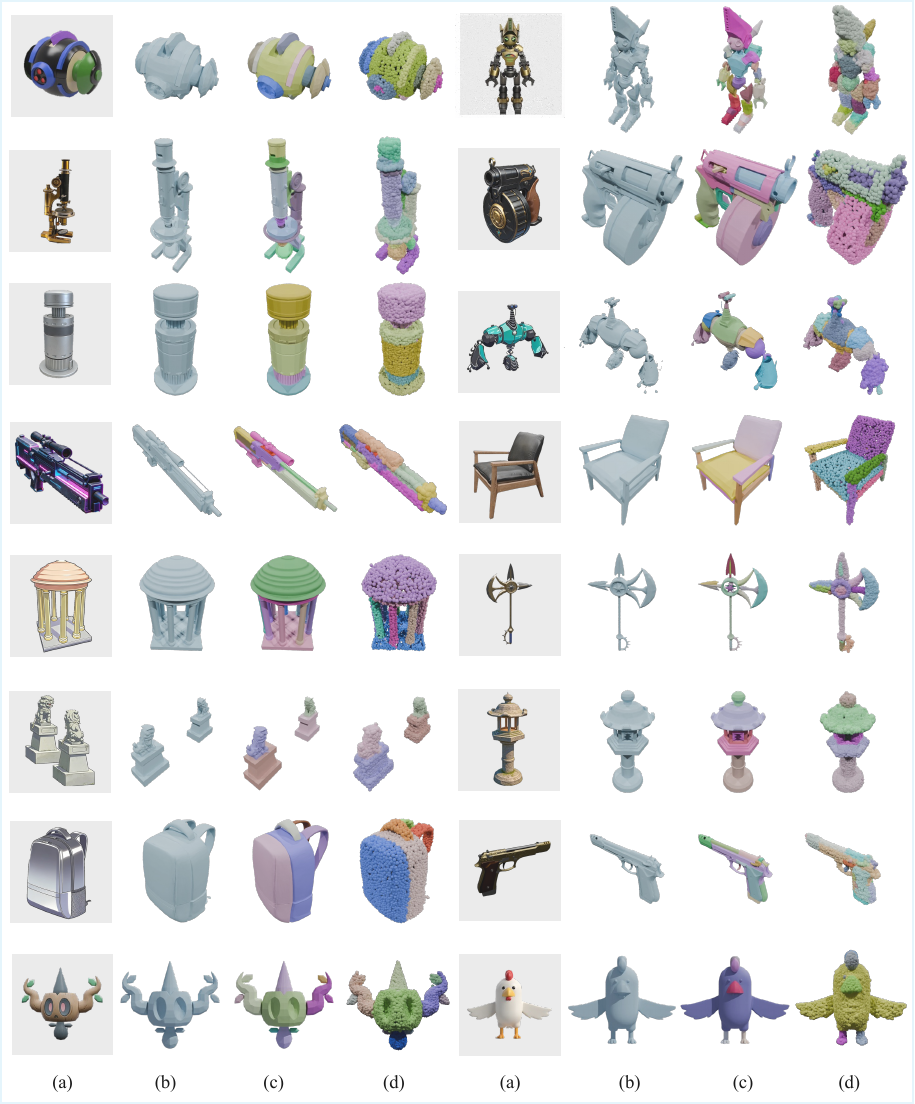}
\caption{Additional generation results using our \methodName. (a) Input images; (b) Whole object geometry generated by our whole-level DiT; (c) Assembled part meshes produced by our part-level DiT; (d) Visualization of part latent segmentation produced by our whole-level DiT. }
\label{fig:more_res_supp_1}
\end{figure*}

\begin{figure*}
\centering
\includegraphics[width=0.9\linewidth]{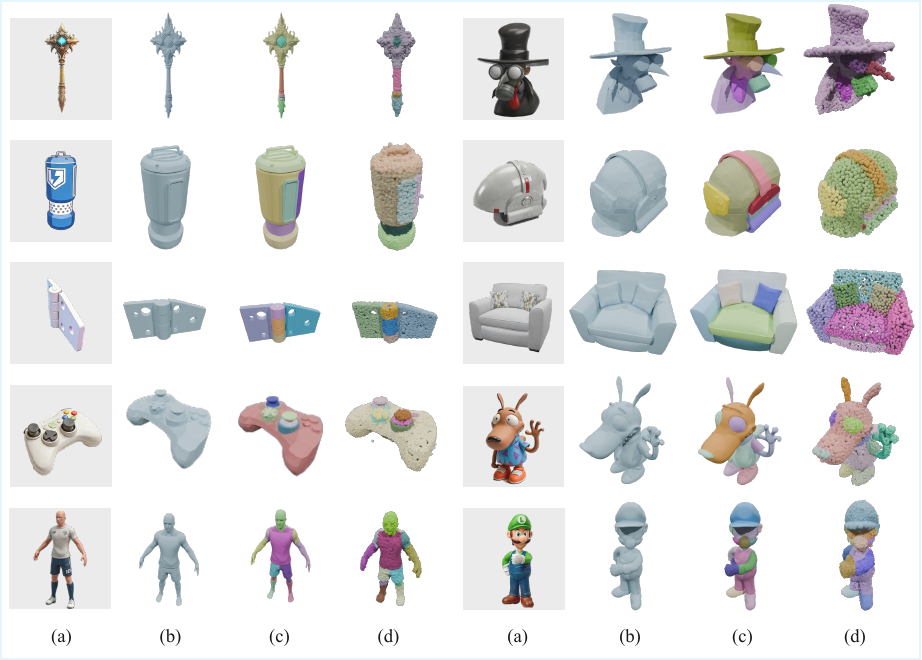}
\caption{More generation results using our \methodName. (a) Input images; (b) Whole object geometry generated by our whole-level DiT; (c) Assembled part meshes produced by our part-level DiT; (d) Visualization of the part latent segmentation produced by our whole-level DiT. }
\vspace{-3mm}
\label{fig:more_res_supp}
\end{figure*}

We provide more generation results of \methodName, including global geometry, part latent segmentation, and part-level geometry, in Fig.~\ref{fig:more_res_supp_1} and Fig.~\ref{fig:more_res_supp} to show the impressive performance of our method and the robustness to various object images.


\end{document}